\documentclass[10pt,onecolumn]{article}
\makeatletter
\DeclareRobustCommand\onedot{\futurelet\@let@token\@onedot}
\def\@onedot{\ifx\@let@token.\else.\null\fi\xspace}

\def\etal{\emph{et al}\onedot}

\def\cl@chapter{\@elt {theorem}}
\makeatother

\usepackage{times}
\usepackage{epsfig}
\usepackage{graphicx}
\usepackage{amssymb}
\usepackage{color}
\usepackage{bm}
\usepackage[utf8]{inputenc} 
\usepackage{caption}
\usepackage{subcaption}
\usepackage{booktabs}
\usepackage{multirow}
\usepackage{epstopdf}
\usepackage{mathtools}
\usepackage{tikz,pgfplots}
\usepackage{footnote}
\usepackage{xspace}
\usepackage{marvosym}
\usepackage{bbm}
\usepackage{geometry}

\newcommand{\Vector}[1]{\mathbf{#1}}
\newcommand{\Matrix}[1]{\mathbf{#1}}

\newcommand{\om}{\bm{\omega}}
\newcommand{\ones}{\bm{\mathbbm{1}}}

\newcommand{\AO}{\boldsymbol{\mathbf{\Omega}}}
\newcommand{\OB}{\mathrm{OB}}
\newcommand{\Mfld}{\mathcal{M}}
\newcommand{\Rmfld}{\mathcal{R}}
\newcommand{\Lfunc}{L}
\newcommand{\TangSp}[2]{T_{#1} #2}
\newcommand{\PllTransp}{\mathrm{\Psi}}
\newcommand{\MessOp}{\mathrm{\Phi}}

\DeclareMathOperator{\tr}{tr}

\newcommand{\envelope}{(\raisebox{-.5pt}{\scalebox{1.45}{\Letter}}\kern-1.7pt)}

\usepackage[colorlinks=true,urlcolor=blue]{hyperref}
\usepackage{cleveref}
\crefformat{footnote}{#2\footnotemark[#1]#3}

\def\abstract
   {%
   \centerline{\large\bf Abstract}%
   \vspace*{12pt}%
   \it%
   }

\begin{document}

\title{A Bimodal Co-Sparse Analysis Model\\
		for Image Processing}
\date{}
\author{Martin Kiechle \and
        Tim Habigt \and
        Simon Hawe \and
        Martin Kleinsteuber\\ \\
        Department of Electrical Engineering and Information Technology,\\
		Technische Universität M\"unchen, Munich, Germany\\
		{\tt\small \{martin.kiechle,tim,simon.hawe,kleinsteuber\}@tum.de} \\
		{\small\url{http://www.gol.ei.tum.de}}
}



\maketitle

\begin{abstract}

The success of many computer vision tasks lies in the ability to exploit the interdependency between different image modalities such as intensity and depth. Fusing corresponding information can be achieved on several levels, and one promising approach is the integration at a low level. Moreover, 
sparse signal models have successfully been used in many vision applications. Within this area of research, the so-called co-sparse analysis model has attracted considerably less attention than its well-known counterpart, the sparse synthesis model, although it has been proven to be very useful in various image processing applications. 

%

In this paper, we propose a co-sparse analysis model that is able to capture the interdependency of two image modalities. It is based on the assumption that a pair of analysis operators exists, so that the co-supports of the corresponding bimodal image structures are 
correlated. We propose an algorithm that is able to learn such a coupled pair of operators from registered and noise-free training data. Furthermore, we explain how this model can be applied to solve linear inverse problems in image processing and how it can be used for image registration tasks. 
This paper extends the work of some of the authors by two major contributions.
Firstly, a modification of the learning process is proposed that a priori guarantees unit norm and zero-mean of the rows of the operator. This accounts for the intuition that contrast in image modalities carries the most information. Secondly, the model is used in a novel bimodal image registration algorithm which estimates the transformation parameters of unregistered images of different modalities.

%
%
\end{abstract}


\section{Introduction}
\label{sec:introduction}
In the past, the majority of methods tackling problems in computer vision were focused on working on a single image modality, typically a color or grayscale image captured with a digital camera. Due to the progress in sensor technologies, sensors that capture different types of image modalities beyond intensity, have become affordable and popular. Well-known examples of multi-modal image sensors include thermal, multispectral and depth cameras, as well as MRI or PET. These image signals often carry information about one another and exploiting this interdependency is beneficial for solving problems in computer vision, such as reconstruction, registration, segmentation, detection, or recognition in a more robust way. Inspired by biological systems which perceive their environment through many different signal modalities at once, fusing sensory information from different modalities has emerged as an important research topic.
Existing fusion schemes can be grouped
according to their level of fusion. Methods of decision-level fusion work independently on the different modalities to make separate task-dependent decisions, which are then fused according to a certain rule or confidence measure. Feature-level fusion methods integrate modality-specific features to derive a decision, for instance the well-known bag-of-words method in object classification. The method presented in this paper, belongs to the group of low-level fusion, where the multimodal information is integrated on the pixel level.

Often, low-level integration is interpreted as finding a mapping from one modality or image domain to another. Typically, this mapping is learned from sets of aligned local image patches to make corresponding algorithms computationally tractable \cite{Freeman2000,Baker2002,HongChang2004,Liu2007}.
More recent approaches aim at capturing the low-level integration across modalities via sparse coding, where the interdependencies of the signals are reflected in interdependencies of their sparse codes.
This concept is used in several methods to find a mapping between different resolution levels or across image modalities. 
In \cite{Yang2010}, Yang \etal apply such a scheme to single image super-resolution (SR). They learn two dictionaries for corresponding low-resolution (LR) and high-resolution (HR) image patches and fuse the two domains through a common sparse representation. In \cite{Li2012}, Li \etal propose a SR approach across the two different image modalities intensity and depth. Three domains are fused through separate dictionaries for LR and HR depth as well as HR intensity patches. The dictionaries are learned
by enforcing common support in the sparse representation.

The assumption of a common sparse code across the different domains 
is often too strict in practice. 
In \cite{Mairal2012}, a less restrictive model is proposed, in which a dictionary is learned for the source image domain together with a transformation matrix which transforms the sparse representations of the source domain to signals in the target domain. Wang \etal \cite{Wang2012} use linear regression between the sparse representations over different dictionaries for the image domains. A similar idea is followed by Jia \etal \cite{Jia2013}, who refine the linear mapping of sparse codes by a local parameter regression for different subsets of sparse representations.
While all of these fusion methods rely on the sparse synthesis model, the related co-sparse analysis model \cite{Elad2007} has not been considered yet in such a multi-modal setting. This is particularly surprising given its excellent performance in unimodal image processing tasks \cite{nam:2011,Hawe2012,Yaghoobi2013}. 

%
%
In this work, we propose a bimodal data model based on co-sparsity for two image modalities. It is based on an extension of the co-sparse analysis model and allows to find signal representations that are simultaneously co-sparse across the two different image domains. We revisit the learning procedure proposed by some of the authors in \cite{kiechle2013jid} with refinements on parameter selection and a modification of the manifold structure on which the model is learned. 
In order to demonstrate both, descriptive power and cross-modal coupling of this model, we first propose to employ it as a prior for solving inverse problems, which we subsequently validate in an image guided depth-map reconstruction task.
The model is further applied in a novel algorithm for bimodal image registration, which, to the best of our knowledge, is the first sparsity-based approach to tackle this problem. Therein, we combine the proposed joint bimodal co-sparsity model with an optimization on Lie groups and achieve favorable results in comparison to other image registration methods.

We outline the remainder of this paper as follows. In Section \ref{sec:model}, the bimodal co-sparse analysis model is described and an appropriate learning objective is derived.
Subsequently, we explain in Section \ref{sec:model_learning} how a solution of this learning objective can be computed efficiently using optimization techniques on matrix manifolds. Sections \ref{sec:reconstruction} and Section \ref{sec:alignment} contain the experiments on bimodal image reconstruction and bimodal image registration. 


\section{Bimodal Co-Sparse Analysis Model}
\label{sec:model}
The (unimodal) co-sparse analysis model \cite{nam:2011} assumes that for a given class of signals $\mathcal{S} \subset \mathbb{R}^n$, there exists a so-called \emph{analysis operator} $\AO \in \mathbb{R}^{k \times n}$ such that the \emph{analyzed vector}
\begin{align}
\AO \mathbf{s} \quad\text{is sparse for all } \mathbf{s}\in \mathcal{S}.
\end{align}

From a geometrical perspective, $\mathcal{S}$ is contained in a un\-ion of subspaces and $\mathbf{s}\in \mathcal{S}$ lies in the intersection of all hyperplanes whose normal vectors are given by the rows of $\AO$ that are indexed by the zero entries of $\AO \mathbf{s}$. This index set is called the \emph{co-support} of $\mathbf{s}$ and is denoted by 
\begin{equation}\label{eq:cosupp}
cosupp(\AO \mathbf{s}):=\{j~|~(\AO \mathbf{s})_j=0\},
\end{equation}
where $(\AO\mathbf{s})_j$ is the $j$-th entry of the analyzed vector.
In image processing applications, $\mathcal{S}$ typically consists of vectorized image patches. One prominent example for a co-sparse analysis model for natural images is the so-called total variation operator. This ad-hoc model assumes that differences of neighboring pixel intensities result in a sparse vector. However, it has been shown that
such ad-hoc models are inferior to models that are adapted to the specific class $\mathcal{S}$ of interest, cf. \cite{Hawe2012,Ravishankar2013a,Yaghoobi2013}. Consequently, \emph{analysis operator learning} aims at finding the most suitable analysis operator for a given class $\mathcal{S}$.

%

In this work, we consider two signal classes $\mathcal{S}_U$ and $\mathcal{S}_V$ of different modalities that emanate from the same physical object. Consider for example an intensity image and a depth map captured from the same scene.
More precisely, let $(\Vector{s}_U, \Vector{s}_V) \in \mathcal{S}_U \times \mathcal{S}_V$. We assume that these signal pairs $(\Vector{s}_U, \Vector{s}_V)$ allow a co-sparse representation with an appropriate pair of analysis operators $(\AO_U, \AO_V)\in \mathbb R^{k \times n_U}\times \mathbb R^{k \times n_V}$. 
%
%
Based on the knowledge that the structure of a signal is encoded in its co-support \eqref{eq:cosupp}, we assume that 
\emph{a pair of analysis operators exists such that the co-supports of $\mathcal{S}_U$ and $\mathcal{S}_V$ are statistically dependent}.
 The bimodal co-sparse analysis model is thus based on the assumption that the conditional probability of $j$ belonging to the co-support of $\Vector{s}_V$ given that $j$ belongs to the co-support of $\Vector{s}_U$ is significantly higher than the unconditional probability, i.e.
\begin{equation}\label{eq:cond_prob}
Pr( \{j \in cosupp(\AO_V \Vector{s}_V) \}~|~\{j \in cosupp(\AO_U \mathbf{s}_U) \}) \gg Pr(\{j \in cosupp(\AO_V \mathbf{s}_V) \}).
\end{equation}
Geometrically interpreted, we aim at partitioning the signal space for each of the two modalities in such a way, that the partitions not only represent subsets of signals of interest but simultaneously relate to a partition of the other modality.

Clearly, this model is idealized, since in practice the entries of the analyzed vectors are not exactly equal to zero. In the following, we relax this strict co-sparsity assumption and show how a coupled pair of analysis operators can be jointly learned, such that aligned bimodal signals analyzed by these operators adhere to the co-sparse model.


More specifically, we aim at learning the coupled pair of bimodal analysis operators $(\AO_U, \AO_V)\in \mathbb{R}^{k\times n_U}\times \mathbb{R}^{k \times n_V}$ for two signal modalities. Therefore, we use a set of $M$ aligned and corresponding training pairs 
\begin{align}
\{(\mathbf{s}_U^{(i)},\mathbf{s}_V^{(i)})\in\mathbb{R}^{n_U} \times \mathbb{R}^{n_V}\}_{i=1}^M.
\end{align}
For simplicity, we assume throughout this work that training signals of both modalities have the same size, i.e\ \hspace{1 mm} $n_U {=} n_V {=} n$.
Now, we incorporate the proposed condition \eqref{eq:cond_prob} into the learning process by inducing the zeros of corresponding analyzed vectors $\AO_U \Vector{s}_{U}^{(i)},\AO_V \Vector{s}_{V}^{(i)}$ to be at the same positions. From here on, the function
\begin{equation}
\Vector{x} \mapsto \sum_{j=1}^k \log(1 + \nu x_j^2),
\end{equation}
with $\nu >0$ being a positive weight and $x_j$ denoting the entries of $\Vector{x}$, serves as an appropriate sparsity measure. Note, that any other smooth sparsity measure principally leads to similar results. With this, the coupled sparsity is controlled through the function
\begin{align}\label{eq:coupling}
g(\AO_U \Vector{s}_U^{(i)},\AO_V \Vector{s}_V^{(i)}) & := \sum \limits_{j=1}^{k} \log \bigg(1 + \nu\big((\AO_U \Vector{s}_{U}^{(i)})_j^2 + (\AO_V \Vector{s}_{V}^{(i)})_j^2\big)\bigg).
\end{align}
Multi-modal dictionary learning methods which model the coupling across modalities by the same sparse representation have found to be too restrictive or require dictionaries of high over-completeness \cite{Jia2013}. Although we also attain the coupling over the sparse representation, there are two key differences. First, the co-sparse analysis model tends to lead to a richer union of subspaces than the synthesis model \cite{nam:2011} and is therefore inherently less restrictive. Second, we relax the strict coupling of a common sparse representation by the smooth function \eqref{eq:coupling}, which less restrictively promotes correlated representations. As it turns out, this results in substantial cross-modal coupling without the need of highly redundant operators.

To find the ideal pair of bimodal operators we minimize the empirical expectation of
\eqref{eq:coupling}
over all training signal pairs, which reads as
\begin{equation}\label{eq:coupling_mat}
G(\AO_U,\AO_V)  := 
\tfrac{1}{M} \sum \limits_{i=1}^{M} g(\AO_U \Vector{s}_U^{(i)},\AO_V \Vector{s}_V^{(i)}).
\end{equation}
%
In order to avoid the trivial solution, some regularization constraints on $\AO$ have to be imposed. In a first step, we demand the rows of $\AO$ to have unit Euclidean norm, i.e. we restrict the transpose of possible solutions to the so-called \emph{oblique manifold} 
\begin{align}
\AO^{\top} \in \OB(n,k):= \mathbb{S}_{n-1}^{\times k},
\end{align}
where $\mathbb{S}_{n-1}$ denotes the unit sphere in $\mathbb{R}^n$.
%
Unfortunately, in general such an operator $\AO$ is biased by signals with varying mean values, i.e.\
\begin{equation}
\AO \Vector{s} \neq \AO (\Vector{s} + c \,\ones_n),
\end{equation}
where $\ones_n$ is the vector with all entries being equal to $1$.
Because we are interested in the structure of an image signal independent of its constant component, we need to avoid the influence of different mean values. For natural images, this can for instance be interpreted as a certain invariance to changes in brightness. A straightforward and popular way to account for such illumination invariance is to learn the model from zero-mean training signals $\, \Vector{s}_c = \Vector{s} - \Vector{\bar{s}} \,$ by subtracting the mean. 
If we denote $\Matrix{I}_n$ as the identity operator and $\Matrix{J}_n$ as a matrix with all elements equal to one, we can express this centering operation in matrix vector notation as
\begin{equation}
\Vector{s}_c = (\Matrix{I}_n - \tfrac{1}{n} \Matrix{J}_n) \, \Vector{s}.
\end{equation}
This operation, however, induces a new trivial solution $\tfrac{1}{\sqrt{n}} \ones_n$ for the rows $\om^{\top}$ of the operator without encoding any structural information.
%
%
Therefore, we extend our approach in \cite{kiechle2013jid} by further restricting the 
rows of 
possible solutions to the orthogonal complement of $\ones_n$, which we shortly denote by $\ones_n^{\perp}$.
Thus, we further restrict the transpose of admissible solutions $\AO$ to 
\begin{equation} \label{eq:intersection_manifold}
\Rmfld = \left(\mathbb{S}_{n-1}\; \cap \; \ones_n^{\perp}\right)^{\times k}.
\end{equation}
Besides these constraints, it has been well investigated \cite{Hawe2012} that coherence and rank are important properties of an analysis operator to well represent a signal class. Therefore we regularize the rank of $\AO|_{\ones_n^{\perp}}$ with the penalty function
\begin{equation}\label{eq:logdet}
h(\AO):= -\tfrac{1}{(n-1)\log(n-1)} \log \det(\tfrac{1}{k} \Matrix{W}^{\top} \AO^\top \AO \Matrix{W}),
\end{equation}
in which the columns of $\Matrix{W} \in \mathbb{R}^{n \times (n-1)}$ form an arbitrary orthonormal basis of $\ones_n^{\perp}$.

Furthermore, to control the coherence of the operator and enforce distinct rows, we adopt the regularizer proposed in \cite{Hawe2012}, which is a log-barrier function on the scalar product of all operator rows, i.e.\
\begin{equation}\label{eq:lindep}
r(\AO):= - \hspace{-4mm} \sum \limits_{1 \leq i < l \leq k} \log(1-({\bm \omega}_{i}^\top {\bm \omega}_{l})^2).
\end{equation}

The combination of the two regularizers
\begin{align}\label{eq:comb}
p(\AO):= \kappa h(\AO) + \mu r(\AO),
\end{align}
with $\kappa,\mu \in \mathbb{R}^+$ being positive weights,
in conjunction with the co-sparsity objective function \eqref{eq:coupling_mat} comprises our final learning function
\begin{equation}\label{eq:learning_func}
\Lfunc \left( \AO_{U}, \AO_{V} \right) := G(\AO_{U},\AO_{V}) + p(\AO_{U}) + p(\AO_{V}).
\end{equation}
Accordingly, the problem of finding the appropriate pair of joint bimodal co-sparse analysis operators is stated as
\begin{equation}\label{eq:BAOproblem}
(\AO_{U}^{\top}, \AO_{V}^{\top}) \in \underset{\Matrix{X}_U ,\Matrix{X}_{V} \in \Rmfld}{\arg\min} \Lfunc(\Matrix{X}^{\top}_{U}, \Matrix{X}^{\top}_{V}).
\end{equation}
\section{Joint Bimodal Analysis Operator Learning Algorithm}
\label{sec:model_learning}

In order to find an optimal solution to \eqref{eq:BAOproblem}, we employ a conjugate gradient method on manifolds.
%
%
To make this work self-contained, we briefly review conjugate gradient and gradient descent methods on matrix manifolds in general and point the interested reader to \cite{abs:oamx08} for further details. Based on this concept, we then derive the learning algorithm for the proposed joint bimodal analysis operators in Section \ref{sec:JBAO_learning}. The proposed optimization framework will also prove useful for the reconstruction and alignment algorithms presented in Section \ref{sec:reconstruction} and Section \ref{sec:alignment}, respectively.
%
\subsection{Line Search Methods on Matrix Manifolds}\label{sec:line_search}
Let $\mathcal{M}$ be a smooth Riemannian sub-manifold of a finite dimensional real vector space $\mathbb{V}$ with a scalar product $\langle \cdot , \cdot \rangle$ and consider the problem of minimizing a smooth real valued function 
\begin{equation} \label{eq:manifold_problem}
f \colon \mathcal{M} \to \mathbb{R}.
\end{equation}
The general idea of line search methods like conjugate gradient or gradient descent algorithms on manifolds is that, starting from some point $\Matrix{X} \in \mathcal{M}$ we search along a curve on the manifold towards the minimizer of \eqref{eq:manifold_problem}.
In our setting, the descent direction is an element of the tangent space $\TangSp{\Matrix{X}}{\Mfld}$, and we search for the updated iterate along geodesics.
In the case where $f$ is defined in the embedding space $\mathbb{V}$, its gradient $\nabla f(\Matrix{X})$ with respect to $\langle \cdot , \cdot \rangle$ is uniquely determined by 
\begin{align}
\tfrac{\rm d}{{\rm d}t}\big|_{t=0} f(\Matrix{X}+ t \Matrix{H}) = \langle \nabla f(\Matrix{X}), \Matrix{H}  \rangle \quad \text{for all } \Matrix{H} \in \mathbb{V}.
\end{align}
The \emph{Riemannian gradient} $\Matrix{G}(\Matrix{X})$, which serves as the (negative) search direction for a gradient descent method on manifolds, is simply the orthogonal projection of $\nabla f(\Matrix{X})$ onto the tangent space $\TangSp{\Matrix{X}}{\Mfld}$, i.e.

%
%
%
%
\begin{equation}\label{eq:riemann_gradient}
\Matrix{G}(\Matrix{X}) = \mathrm{\Pi}_{\TangSp{\Matrix{X}}{\Mfld}} \big( \nabla f(\Matrix{X}) \big),
\end{equation}
with $\mathrm{\Pi}_{\TangSp{\Matrix{X}}{\Mfld}}$ denoting the orthogonal projection with respect to $\langle \cdot , \cdot \rangle$.
Now let $t \mapsto \mathrm{\Gamma} (\Matrix{X}, \Matrix{H}, t)$ denote the geodesic emanating from $\Matrix{X} \in \Mfld$ in direction $\Matrix{H} \in \TangSp{\Matrix{X}}{\Mfld}$, that is 
\begin{align}
\mathrm{\Gamma} (\Matrix{X}, \Matrix{H}, 0)= \Matrix{X} \quad\text{and}\quad 
\tfrac{\rm d}{{\rm d}t}\big|_{t=0} \mathrm{\Gamma} (\Matrix{X}, \Matrix{H}, t) = \Matrix{H}.
\end{align}
%
Schematically, line search methods on manifolds update the $i$-th estimate $\Matrix{X}^{i}$ by
\begin{equation}\label{eq:est_update}
\Matrix{X}^{i+1} = \mathrm{\Gamma} (\Matrix{X}^i, \Matrix{H}^i, t^i),
\end{equation}
where $\Matrix{H}^i \in \TangSp{\Matrix{X}}{\Mfld}$ is the descent direction and $t^i \in \mathbb{R}$ is a suitable step-size.
If $\Matrix{H}^i= - \Matrix{G}(\Matrix{X}^{i})$ is the negative Riemannian gradient, the method is a gradient descent algorithm and provides linear convergence to a local minimizer for appropriate step-size selections \cite{abs:oamx08}.

%

In practice, faster convergence can often be achieved by adapting conjugate gradient methods to the manifold setting. In this case, the search direction $\Matrix{H}^{i+1} \in \TangSp{\Matrix{X}^{i+1}}{\Mfld}$ is a linear combination of the Riemannian gradient $\Matrix{G}^{i+1}:=\Matrix{G}(\Matrix{X}^{i+1}) \in \TangSp{\Matrix{X}^{i+1}}{\Mfld}$ and the previous search direction $\Matrix{H}^{i}$. Since the linear combinations of elements from different tangent spaces are not defined, the parallel transport along geodesics is employed to identify the different tangent spaces. 
If we denote this parallel transport by
\begin{align}
\PllTransp_{i}^{i+1} \colon \TangSp{\Matrix{X}^{i}}{\Mfld} \to \TangSp{\Matrix{X}^{i+1}}{\Mfld},
\end{align}
the conjugate gradient method on manifold updates the search direction via

\begin{equation}
\Matrix{H}^{i+1} := -\Matrix{G}^{i+1} + \beta^i \, \PllTransp_{i}^{i+1}(\Matrix{H}^i),
\end{equation}
where initially, $\Matrix{H}^{0}:= -\Matrix{G}^{0}$.
For our purposes, the update parameter $\beta^i$ is chosen according to a manifold adaption of the
Fletcher-Reeves and Dai-Yuan formula. More precisely, 
 we employ a hybridization of the Hestenes-Stiefel Formula and the Dai Yuan formula 
\begin{align}\label{eq:dyhs}
\beta^{(i)}_{\textit{hyb}} = \max\big (0,\min(\beta^{(i)}_{\textit{DY}},\beta^{(i)}_{\textit{HS}})\big),
\end{align}
which has been suggested in \cite{cg:dai:2001}, where
\begin{align}
\beta^{(i)}_{\textit{HS}} & = 
 \frac{\langle\Matrix{G}^{(i+1)},\Matrix{G}^{i+1}-\mathrm{\Psi}_i^{i+1}(\Matrix{G}^{i}) \rangle}{\langle \mathrm{\Psi}_i^{i+1}(\Matrix{H}^{i}),\Matrix{G}^{i+1}-\mathrm{\Psi}_i^{i+1}(\Matrix{G}^{i})\rangle},\label{eq:hs}\\
\beta^{(i)}_{\textit{DY}} & =  \frac{\langle \Matrix{G}^{(i+1)},\Matrix{G}^{(i+1)}\rangle}{\langle \mathrm{\Psi}_i^{i+1}(\Matrix{H}^{i}),\Matrix{G}^{i+1}-\mathrm{\Psi}_i^{i+1}(\Matrix{G}^{i})\rangle}\label{eq:dy}.
\end{align}

%
%

%
\subsection{Geometric Conjugate Gradient for Joint Bimodal Analysis Operator Learning} \label{sec:JBAO_learning}


We propose a geometric conjugate gradient method as explained in the previous subsection to tackle problem \eqref{eq:BAOproblem} for learning the joint bimodal analysis operator. First, we have to ensure that 
$\Rmfld$ as given in \eqref{eq:intersection_manifold} is indeed a manifold.

\emph{Lemma.} The set $\Rmfld = \left(\mathbb{S}_{n-1} \cap  \ones_n^{\perp}\right)^{\times k}$ is a Riemannian submanifold of $\mathbb{R}^{n \times k}$ and the tangent space at $\Matrix{X}=[\Vector{x}_1, \dots \Vector{x}_k]\in \Rmfld$ is given by
\begin{align}
\TangSp{\Matrix{X}}{\Rmfld} =\TangSp{\Vector{x}_1}{(\mathbb{S}_{n-1} \cap  \ones_n^{\perp})} \times \dots \times 
\TangSp{\Vector{x}_k}{(\mathbb{S}_{n-1} \cap  \ones_n^{\perp})},
\end{align}
with
\begin{align}\label{eq:tangent_small}
\TangSp{\Vector{x}}{(\mathbb{S}_{n-1} \cap  \ones_n^{\perp})}=\left\{\Vector{h} \in \mathbb{R}^n ~|~\Vector{h}^\top [\Vector{x}, \ones_n] = 0 \right\}.
\end{align}

\emph{Proof.} By using the product manifold structure, it is sufficient to show that $\mathbb{S}_{n-1}\; \cap \; \ones_n^{\perp}$ is a submanifold of $\mathbb{R}^n$ with its tangent space as given in \eqref{eq:tangent_small}.
Consider the function
\begin{align}
F\colon \mathbb{R}^n \to \mathbb{R}^2, \quad \Vector{x} \mapsto \begin{bmatrix} \|\Vector{x}\|^2-1 \\ \Vector{x}^\top \ones_n \end{bmatrix}.
\end{align}
Then $\mathbb{S}_{n-1}\cap  \ones_n^{\perp}=F^{-1}(0)$ and the derivative of $F$ is given by
\begin{align}
DF(\Vector{x}) \Vector{h}= \begin{bmatrix} 2 \Vector{x}^\top \\ \ones_n^\top \end{bmatrix} \Vector{h},
\end{align}
which is surjective for all $\Vector{x} \in \mathbb{S}_{n-1}\cap  \ones_n^{\perp}$.
The regular value theorem now implies that $\mathbb{S}_{n-1} \cap  \ones_n^{\perp}$ is a submanifold of $\mathbb{R}^n$ and that $\TangSp{\Vector{x}}{(\mathbb{S}_{n-1} \cap  \ones_n^{\perp})}$ is given by the null space of $DF(\Vector{x})$, yielding equation \eqref{eq:tangent_small}. \hfill{$\Box$}

With respect to the standard inner product, the orthogonal projection onto $\TangSp{\Vector{x}}{(\mathbb{S}_{n-1} \cap  \ones_n^{\perp})}$ is given by the projection matrix
\begin{align} \label{eq:orthogonal_projector}
\Matrix{P}_\Vector{x}  = \left(\Matrix{I}_n - \Matrix{Q}_\Vector{x}\Matrix{Q}_\Vector{x}^{\top} \right),
\end{align}
where  
\begin{equation}
\Matrix{Q}_\Vector{x}= \left[ \Vector{x}, \tfrac{1}{\sqrt{n}} \ones_n \right] 
\end{equation}
has orthonormal columns.
Using the product manifold structure, we find the orthogonal projection from $\mathbb{R}^{k \times n}$ onto $\TangSp{\Matrix{X}}{\Rmfld}$ as
\begin{align}\label{eq:projection_final}
\mathrm{\Pi}_\Matrix{X} \left[\Vector{y}_1, \dots, \Vector{y}_k \right] = \left[\Matrix{P}_{\Vector{x}_1} \Vector{y}_1, \dots, \Matrix{P}_{\Vector{x}_k} \Vector{y}_k \right].
\end{align}

In order to compute the Riemannian gradient of the learning function \eqref{eq:learning_func}, note that the gradient with respect to the standard inner product is given by
\begin{align}
\nabla \Lfunc (\Matrix{X}_U^\top, \Matrix{X}_V^\top)=[\nabla_U \Lfunc (\Matrix{X}_U^\top, \Matrix{X}_V^\top)^\top, \nabla_V \Lfunc (\Matrix{X}_U^\top, \Matrix{X}_V^\top)^\top],
\end{align}
where $\nabla_U$ and $\nabla_V$ denote the gradient of $\Lfunc$ with respect to its first and second input. Using equation \eqref{eq:projection_final}, the Riemannian gradient is thus
\begin{align}
\Matrix{G}& (\Matrix{X}_U, \Matrix{X}_V) = [\mathrm{\Pi}_{\Matrix{X}_U} \nabla_U \Lfunc (\Matrix{X}_U^\top, \Matrix{X}_V^\top)^\top, \mathrm{\Pi}_{\Matrix{X}_V} \nabla_V \Lfunc (\Matrix{X}_U^\top, \Matrix{X}_V^\top)^\top].
\end{align}

Since $\Rmfld$ is a submanifold of the oblique manifold $\OB(n,k)$, the formulas for the geodesics and the parallel transport coincide. We refer to \cite{Hawe2012} for explicit formulas for the geodesics and the parallel transport. Following the general conjugate gradient scheme presented in subsection \ref{sec:line_search}, it is now straightforward to implement the learning algorithm.

%
%

Concerning the choice of training samples, we randomly select $M$ pairs of aligned patches $(\Vector{s}_U^{(i)}, \Vector{s}_V^{(i)})$ from noise-free images. Since typically, the two modalities are measured in different physical units, the patches are normalized by their standard deviation to allow a comparison of both modalities.  Patches with small standard deviation are discarded for the learning process. Note, due to the restriction of the admissible set of operators to
$\Rmfld$, cf. \eqref{eq:intersection_manifold},
 patches with small standard deviation (i.e. nearly constant patches) generically fit our model and thus discarding them does not bias the learning process.

%
%

\begin{figure}
	\centering
	\begin{subfigure}[b]{0.48\columnwidth}
		\centering
		\includegraphics[width=\textwidth]{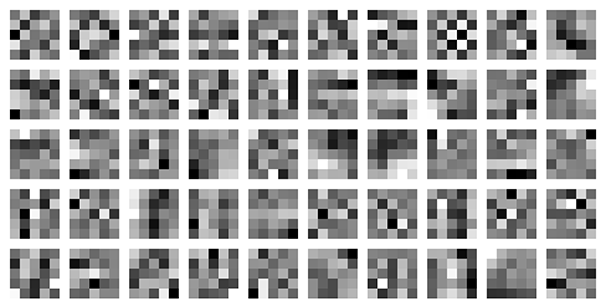}
		\caption{intensity}
		\label{fig:AOP_ID_I}
	\end{subfigure}%
	\,
	\begin{subfigure}[b]{0.48\columnwidth}
		\centering
		\includegraphics[width=\textwidth]{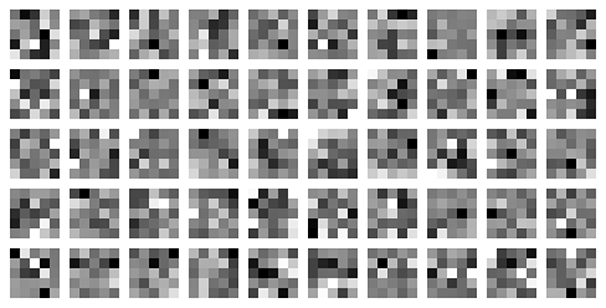}
		\caption{depth}
		\label{fig:AOP_ID_D}
	\end{subfigure}%
	\medskip
	
	\begin{subfigure}[b]{0.48\columnwidth}
		\centering
		\includegraphics[width=\textwidth]{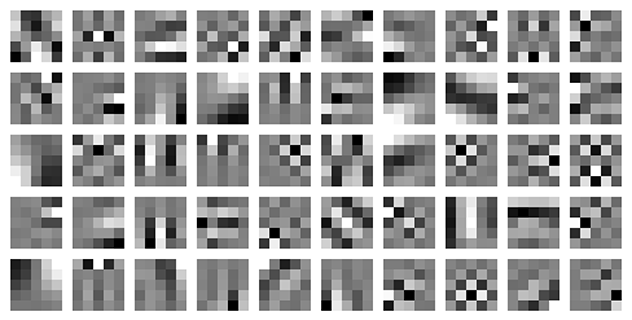}
		\caption{intensity}
		\label{fig:AOP_IN_I}
	\end{subfigure}%
	\,
	\begin{subfigure}[b]{0.48\columnwidth}
		\centering
		\includegraphics[width=\textwidth]{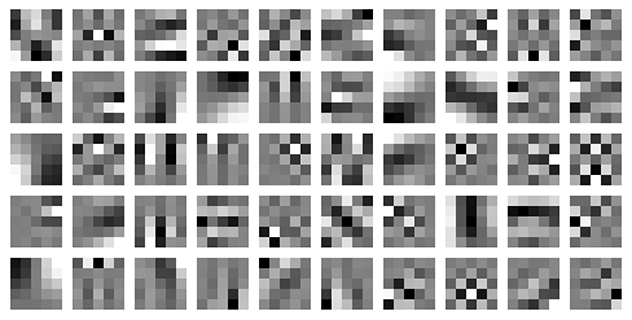}
		\caption{near infrared}
		\label{fig:AOP_IN_N}
	\end{subfigure}%
	\caption{Plot of the rows of bimodal operator pairs visualized as square patches for intensity and depth (top row) as well as intensity and near infrared (bottom row). Patches at the same positions within their plots correspond to rows in the operators with the same index.}\label{fig:operator_atoms}
\end{figure}

Figure \ref{fig:operator_atoms} illustrates the rows of learned operator pairs as square patches for the two bimodal image setups intensity and depth as well as intensity and near infrared (NIR). These are used in the experiments in Section \ref{sec:middlebury_evaluation} and Section \ref{sec:alignment_experiment}. Note how the intensity operators differ between the two setups due to the bimodal coupling with depth and infrared modalities respectively.

Learning such pairs of operators from several thousand samples on a standard desktop PC can be accomplished within the order of a few minutes.

%
%
%
%

\section{Bimodal Image Reconstruction}
\label{sec:reconstruction}
The proposed model is based on learning from local patches. Yet, we want to apply this model to entire images of much larger dimension. Therefore, we first need a global formulation of the local operators to process images. Such a formulation is developed in Section \ref{sec:global_operator}. In Section \ref{sec:image_recon_problem}, we will then show how image reconstruction can be achieved using these global operators and demonstrate its practical applicability in a super-resolution experiment in Section \ref{sec:middlebury_evaluation}.

\subsection{Application of the Patch-Based Operators to Images} \label{sec:global_operator}
According to \cite{Hawe2012}, a \emph{global} analysis operator $\AO^{F} \in \mathbb{R}^{K \times N}$ is constructed from a patch based operator $\AO \in \mathbb{R}^{k \times n}$ as follows. Denote the operator that extracts the
normalized
($\sqrt{n} \times \sqrt{n}$)-dimensional patch centered at position $(r,c)$ from the entire image as $\mathcal{P}_{rc} \in \mathbb{R}^{n \times N}$. The global analysis operator is then given as
\begin{equation} \label{eq:op}
\AO^{F} := 
{
\begin{bmatrix} 
\AO\mathcal{P}_{11} \\ 
\AO\mathcal{P}_{21} \\ 
\vdots 																							\\ 
\AO\mathcal{P}_{hw} 
\end{bmatrix}
}
\in \mathbb{R}^{K \times N},
\end{equation}
with
$K = Nk$, i.e.\ all patch positions are considered. We use the reflective boundary condition to deal with areas along image borders.

\subsection{Formulation of the Bimodal Image Reconstruction Problem} \label{sec:image_recon_problem}
The general goal of the bimodal image reconstruction task is to recover an aligned pair of bimodal images $\mathbf{s}_U,\mathbf{s}_V \in \mathbb{R}^N$ from a set of measurements  $\mathbf{y}_U \in \mathbb{R}^{m_U},\mathbf{y}_V \in \mathbb{R}^{m_V}$.
Here, $\mathbf{s}_U,\mathbf{s}_V$ are the vectorized versions of the original images from each of the two modalities, obtained by ordering their entries lexicographically. Furthermore, $w$, $h$ denote the height and width of both original images and $N {=} wh\,$.

In our reconstruction approach, we treat the problem of bimodal image reconstruction as a linear inverse problem. Formally, the relation between $\mathbf{s}_U, \mathbf{s}_V$ and $\mathbf{y}_U, \mathbf{y}_V$ is given by
%
\begin{equation}\label{eq:inv_problem}
\mathbf{y}_U = \MessOp_U\mathbf{s}_U+\mathbf{e}_U, \qquad \mathbf{y}_V = \MessOp_V\mathbf{s}_V+\mathbf{e}_V.
\end{equation}
$\MessOp_U\in\mathbb{R}^{m_U \times N}, \MessOp_V\in\mathbb{R}^{m_V \times N}$ model the sampling process of the measurements and $\mathbf{e}_U\in\mathbb{R}^{m_U}, \mathbf{e}_V \in\mathbb{R}^{m_V}$ model noise and potential sampling errors. For typical reconstruction tasks, the dimensions $m_U, m_V$ of the measurement vectors may be significantly smaller than the dimension $N$. Consequently, reconstructing $\mathbf{s}_U, \mathbf{s}_V$ in \eqref{eq:inv_problem} is highly ill-posed.

To resolve this, the bimodal data model is employed as a co-sparsity prior to regularize the image reconstruction. Accordingly, we aim to solve
\begin{equation}
\begin{split}
& (\mathbf{s}^\star_U,\mathbf{s}^\star_V) \in \arg \underset{\mathbf{s}_U,\mathbf{s}_V\in \mathbb{R}^{N}} {\min} g(\AO_U^F \mathbf{s}_U,\AO_V^F\mathbf{s}_V) \label{eq:image_reconstruction_fullimage}\\
& \text{subject to} \quad d_{E} \left( \left( \MessOp_U\mathbf{s}_U, \MessOp_V\mathbf{s}_V \right), \left( \mathbf{y}_U, \mathbf{y}_V \right) \right)  \leq \varepsilon. 
\end{split}
\end{equation}
We denote $d_{E}$ as an appropriate data fidelity measure such as the squared Euclidean distance and $\varepsilon \in \mathbb{R}^{+}_0$ is an estimated upper bound of the noise energy.
%
%
%
Consequently, the analyzed versions of both modalities are enforced to have a correlated co-support and as a result, the two signals are coupled. 

Depending on the choice of the measurement operators $\MessOp_U, \MessOp_V$, different reconstruction tasks such as denoising, inpainting, or upsampling can be performed. This can be accomplished jointly on both signals simultaneously or only on one single modality, while the other reinforces the co-sparsity and data priors. We show in the following section how image guided depth map SR can be accomplished using this model.

\subsection{Image-guided Depth Map Super-Resolution}
\label{sec:middlebury_evaluation}
In this experiment, we apply the proposed reconstruction approach to the image modalities intensity and depth. Due to the availability of affordable sensors, this has become a common bimodal image setup. We now focus on recovering the HR depth image $\mathbf{s}_D$ from LR depth measurements $\mathbf{y}_D$, given a fixed high quality intensity image $\mathbf{s}_I{=}\mathbf{y}_I$. In this case, $\MessOp_I$ is the identity operator 
and the analyzed intensity image is constant, i.e.\
\begin{equation}
\AO^{F}_I\mathbf{s}_I=\mathbf{c}=\textit{const.}
\end{equation}
This simplifies problem \eqref{eq:image_reconstruction_fullimage} for recovering the HR depth map to
\begin{equation}
\begin{split}
& \mathbf{s}^\star_D \in \arg\underset{\mathbf{s}_D\in \mathbb{R}^{N}} {\min} \, g(\mathbf{c},\AO^{F}_D\mathbf{s}_D) \label{eq:depth_reconstruction}\\
& \text{subject to} \quad d_{E}(\MessOp_D\mathbf{s}_D,\mathbf{y}_D) \leq \varepsilon_{D}.
\end{split}
\end{equation}
The data fidelity term $d_{E}$ depends on the error model of the depth data and can be chosen accordingly. For instance, this may be an error measure tailored to a sensor specific error model, as described for the Kinect sensor in \cite{kiechle2013jid}.
%
In this way, knowledge about the scene gained from the intensity image and its co-support regarding the bimodal analysis operators helps to determine the HR depth signal.
To compare our results to state-of-the-art methods, we quantitatively evaluate our algorithm on the four standard test images 'Tsukuba', 'Venus', 'Teddy', and 'Cones' from the Middlebury dataset \cite{Scharsteina}. To artificially create LR input depth maps, we scale the ground truth depth maps down by a factor of $d$ in both vertical and horizontal dimension. We first blur the available HR image with a Gaussian kernel of size $(2d-1)\times(2d-1)$ and standard deviation $\sigma=d/3$ before downsampling. The LR depth map and the corresponding HR intensity image are the input to our algorithm.

\begin{savenotes}

	\begin{table}[tp]
	  \centering
	  \resizebox{.5\linewidth}{!}{
		  \begin{tabular}{rrcccc}
		    \toprule
		        $d$ & method & Tsukuba & Venus & Teddy & Cones \\
		    \midrule
		    \multicolumn{1}{c}{\multirow{4}[0]{*}{2x}} & nearest-neighbor & 1.24 & 0.37	& 4.97 & 2.51 \\
		    \multicolumn{1}{c}{} & Yang \etal \cite{Yang2007} & 1.16  & 0.25  & 2.43  & \underline{2.39} \\  
		    \multicolumn{1}{c}{} & Diebel \etal \cite{Diebel2005} & 2.51  & 0.57  & 2.78  & 3.55 \\              
		    \multicolumn{1}{c}{} & Hawe \etal \cite{Hawe2012}\footnote{\label{first}only takes depth as input and therefore solves a harder problem} & 1.03 & 0.22 & 2.95 & 3.56 \\
		    \multicolumn{1}{c}{} & JID \cite{kiechle2013jid} & \textbf{0.47} & \textbf{0.09} & \textbf{1.41} & \textbf{1.81} \\
		    \multicolumn{1}{c}{} & our method  & \underline{0.83} & \underline{0.12} & \underline{1.96} & {2.69} \\
		    \midrule
		    \multicolumn{1}{c}{\multirow{4}[0]{*}{4x}} & nearest-neighbor & 3.53 & 0.81	& 6.71 & 5.44 \\ 
		    \multicolumn{1}{c}{} & Yang \etal & 2.56  & 0.42  & 5.95  & \underline{4.76} \\
		    \multicolumn{1}{c}{} & Diebel \etal & 5.12  & 1.24  & 8.33  & 7.52 \\
		    \multicolumn{1}{c}{} & Hawe \etal & 2.95  & 0.65  & 4.80  & 6.54 \\
		    \multicolumn{1}{c}{} & JID & \underline{1.73} & \underline{0.25} & \textbf{3.54} & 5.16 \\
		    \multicolumn{1}{c}{} & our method  & \textbf{1.48} & \textbf{0.23} & \underline{3.99} & \textbf{4.69} \\
		    \midrule
		    \multicolumn{1}{c}{\multirow{5}[0]{*}{8x}} & nearest-neighbor & 3.56 & 1.90 & 10.9 & 10.4 \\
		    \multicolumn{1}{c}{} & Yang \etal & 6.95  & 1.19  & 11.50 & 11.00 \\   
		    \multicolumn{1}{c}{} & Diebel \etal & 9.68  & 2.69  & 14.5  & 14.4 \\
		    \multicolumn{1}{c}{} & Lu \etal \cite{Lu2011a} & 5.09  & 1.00  & 9.87  & 11.30 \\
		    \multicolumn{1}{c}{} & Hawe \etal & 5.59  & 1.24  & 11.40 & 12.30 \\
		    \multicolumn{1}{c}{} & JID & \underline{3.53} & \textbf{0.33} & \textbf{6.49} & \underline{9.22} \\
		    \multicolumn{1}{c}{} & our method  & \textbf{3.30} & \underline{0.34} & \underline{8.11} & \textbf{8.57} \\
		    \bottomrule
		  \end{tabular}%
	  }
	  \caption{Numerical comparison of our method to other depth map SR approaches for different upscaling factors $d$. The figures represent the percentage of bad pixels with respect to all pixels of the ground truth data and an error threshold of $\delta=1$. Bold and underlined figures highlight the best and second best results.}
	  \label{tab:middlebury_results}%
	  \vspace{10mm}
	\end{table}%

	\begin{table}[tp]
	  \centering
	  \resizebox{.5\linewidth}{!}{
		  \begin{tabular}{rrcccc}
		    \toprule
					$d$ & method & Tsukuba & Venus & Teddy & Cones \\
		    \midrule
		    \multicolumn{1}{c}{\multirow{5}[0]{*}{2x}} & nearest-neighbor & 0.612 & 0.288 & 1.543 & 1.531\\
				\multicolumn{1}{c}{} & Chan \etal \cite{Chan2008} & n/a & 0.216 & 1.023 & 1.353\\   
		    \multicolumn{1}{c}{} & Hawe \etal \cite{Hawe2012}\cref{first}& 0.278 & 0.105 & 0.996  & 0.939 \\
		    \multicolumn{1}{c}{} & JID \cite{kiechle2013jid} & \textbf{0.255} & \textbf{0.075} & \textbf{0.702} & \textbf{0.680}\\
		    \multicolumn{1}{c}{} & our method & \underline{0.256} & \underline{0.077} & \underline{0.803} & \underline{0.821}\\
		    \midrule
		    \multicolumn{1}{c}{\multirow{5}[0]{*}{4x}} & nearest-neighbor & 1.189 & 0.408 & 1.943 & 2.470 \\
				\multicolumn{1}{c}{} & Chan \etal & n/a & 0.273 & \textbf{1.125} & 1.450 \\
		    \multicolumn{1}{c}{} & Hawe \etal & \underline{0.450} & 0.179 & 1.389 & 1.398 \\
		    \multicolumn{1}{c}{} & JID & 0.487 & \underline{0.129} & 1.347 & \underline{1.383} \\
		    \multicolumn{1}{c}{} & our method & \textbf{0.374} & \textbf{0.108} & {1.256} & \textbf{1.287}\\
		    \midrule
		    \multicolumn{1}{c}{\multirow{4}[0]{*}{8x}} & nearest-neighbor & 1.135 & 0.546 & 2.614 & 3.260 \\
				\multicolumn{1}{c}{} & Chan \etal & n/a & 0.369 & \textbf{1.410} & \textbf{1.635} \\
		    \multicolumn{1}{c}{} & Hawe \etal & \underline{0.713} & 0.249 & 1.743 & 1.883 \\
		    \multicolumn{1}{c}{} & JID & 0.753 & \underline{0.156} & 1.662 & \underline{1.871} \\
		    \multicolumn{1}{c}{} & our method & \textbf{0.660} & \textbf{0.155} & \underline{1.729} & {1.931}\\
		    \bottomrule
		  \end{tabular}%
	  }
	\caption{Numerical comparison of our method to other depth map SR approaches. The figures represent the RMSE in comparison with the ground truth depth map. Bold and underlined figures highlight the best and second best results.}
	\label{tab:rmse_results}%
	\end{table}%

In this reconstruction from LR measurements, we assume an i.i.d.\ normal distribution of the error, which leads to
the data fidelity term
\begin{equation}
d_{E}(\MessOp_D\mathbf{s}_D,\mathbf{y}_D)=\Vert\MessOp_D\mathbf{s}_D-\mathbf{y}_D\Vert ^2_2.
\end{equation}
Finally, we obtain the unconstrained formulation of problem \eqref{eq:depth_reconstruction} for reconstructing the HR depth image, namely
\begin{equation}
\mathbf{s}^\star_D \in \underset{\mathbf{s}_D\in \mathbb{R}^{N}} {\arg\min} \, \lambda g(\mathbf{c},\AO^{F}_D\mathbf{s}_D) + \Vert\MessOp_D\mathbf{s}_D-\mathbf{y}_D\Vert ^2_2.
\label{eq:depth_reconstruction_middlebury}
\end{equation}
We solve problem \eqref{eq:depth_reconstruction_middlebury} using a standard conjugate gradient method and an Armijo step size selection.
In that matter, larger values of the weighting factor
$\lambda$
lead to a faster convergence of the algorithm but allow larger differences between the measurements and the reconstruction estimate.
To achieve the best results within few iterations, we start with a large value of $\lambda$ and restart the conjugate gradient optimization procedure several times, while consecutively shrinking the multiplier to a final value of $\lambda=1$.
Problem \eqref{eq:depth_reconstruction_middlebury} is not convex and convergence to a global minimum can not be guaranteed. In practice however, we observe convergence to accurate depth maps from random initializations of $\Vector{s}_D$.
For the evaluation of our approach, we train one fixed operator pair and use it in all presented intensity and depth experiments. To that end, we
gather a total of $M=15000$ pairs of square sample patches of size $\sqrt{n}=5$ from the five registered intensity and depth image pairs 'Baby1', 'Bowling1', 'Moebius', 'Reindeer' and 'Sawtooth' of the Middlebury stereo set \cite{Scharsteina}.
Furthermore, we learn the operators with twofold redundancy, i.e.\ $k=2n$, resulting in the operator pair $(\AO_{I}, \AO_{D}) \in \mathbb{R}^{50 \times 25} \times \mathbb{R}^{50 \times 25}$. In general, a larger redundancy of the operators leads to better reconstruction quality but at the cost of an increased computational burden of both learning and reconstruction. Twofold redundancy provides a good trade-off between reconstruction quality and computation time. We empirically set the remaining learning parameters to $\nu=400, \kappa_I=5, \kappa_D=22, \mu_I=10^2$ and $\mu_D=2.5\cdot 10^4$.

Following the methodology described in the work of comparable depth map SR approaches, we use the Middlebury stereo matching online evaluation tool\footnote{http://vision.middlebury.edu/stereo/eval/} \end{savenotes}to quantitatively assess the accuracy of our results with respect to the ground truth data. We report the percentage of bad pixels over all pixels in the depth map with an error threshold of $\delta=1$. Additionally, we provide the root-mean-square error (RMSE) based on 8-bit images.
We compare our results to several of the state-of-the-art methods for image guided depth map SR.
Here, we focus on methods that conduct the same experiments and point the reader to \cite{kiechle2013jid} for a more comprehensive review of related methods. 

In \cite{Yang2007}, Yang \etal apply bilateral filtering to depth cost volumes in order to iteratively refine an estimate using an additional color image. Chan \etal \cite{Chan2008} elaborate on this approach with a fast and noise-aware joint bilateral filter. In the work of \cite{Diebel2005}, color image information is used to guide depth reconstruction by computing the smoothness term in Markov-Random-Field formulation according to texture \linebreak derivatives, which is extended in \cite{Lu2011a} by a data term better adapted to depth images. We also compare our results to a unimodal co-sparse analysis operator proposed by some of the authors in \cite{Hawe2012}, which we learn from depth samples only. Since the unimodal approach has to solve a harder problem, this demonstrates how a bimodal approach can practically contribute to improvements in reconstruction quality. For completeness, we also include the results achieved with the joint intensity and depth (JID) method proposed earlier \cite{kiechle2013jid}.
Where an implementation is not publicly available, we rely on the results reported by the respective authors regarding the numerical comparison in Table \ref{tab:middlebury_results} and Table \ref{tab:rmse_results}.

Our method improves depth map SR considerably over simple interpolation approaches. Neither staircasing nor substantial blurring artifacts occur, particularly in areas with discontinuities. Also, there is no noticeable texture cross-talk in areas of smooth depth and cluttered intensity. Edges can be preserved with great detail due to the additional knowledge provided by the intensity image, even if SR is conducted using large upscaling factors. The quantitative comparison with other depth map SR methods demonstrates the excellent performance of our approach. The improvement over other methods is of particular significance for larger magnification factors.


\section{Bimodal Image Registration}
\label{sec:alignment}
Image registration is the process of geometrically aligning two images that were taken by e.g. different sensors, at different points in time or from different viewpoints.
Automatic image registration can be categorized into feature-based and area-based algorithms. The first group of algorithms searches for salient features in both images (e.g. edges, corners, contours) and tries to find the matching pairs of features. The geometric transformation that minimizes the distance between matching features is then used to transform one of the images. Area-based algorithms do not consider at specific features but use the whole overlapping region between both images to evaluate the registration.
In both cases a distance metric is needed to either match the features or to measure the similarity between image regions.
In the unimodal registration case simple metrics like the sum of squared differences or correlation can be used.
Multimodal registration is more challenging because the intensities of two different sensors can differ substantially when imaging the same physical object. This phenomenon is often called contrast reversal, as bright objects in one modality can be very dark in the other and vice versa. In general, no straight-forward functional relationship between the intensities of the sensors exists. Nevertheless, the approach of Orchard \cite{Orchard2007} tries to find a piecewise linear mapping between the intensities of different modalities.
The most popular metric for multimodal image registration is Mutual Information, originally introduced by Viola and Wells \cite{Viola1997} and Collignon \etal \cite{Collignon1995}. A normalized version was later proposed by Studholme \etal \cite{Studholme1999} that is better suited to changing sizes of the overlapping region.
Mutual Information is used for a variety of different applications and sensors as in medical registration \cite{Pluim2003}, remote sensing \cite{Cole-Rhodes2003,Fan2010} and surveillance \cite{Krotosky2007}.
For more information about image registration, we refer to Brown \cite{Brown1992} and Zitová and Flusser \cite{Zitova2003} who have published excellent overviews covering several decades of research in this area.

\subsection{Bimodal Image Registration Algorithm}
\label{sec:registrationAlgorithm}
In this section we present an area-based approach that employs the formerly learned bimodal co-sparse analysis model for registration of two image modalities. We consider two images $I_U$ and $I_V$ of a 3D scene that are sensed through two modalities $U$ and $V$. We further assume that these images can be aligned with a transformation $\tau$ that belongs to one of the following Lie groups $\mathcal{G}$.
\begin{itemize}
\item[$\bullet$] The special orthogonal group $SO(2)$;
\item[$\bullet$] the special Euclidean group $SE(2)$;
\item[$\bullet$] the special affine group $SA(2)$;
\item[$\bullet$] or, the affine group $A(2)$.
\end{itemize}

This means that, if $\Vector{x}$ denotes the homogeneous pixel coordinates for one modality, say $I_U$, there exists some $\tau \in \mathcal{G}$ such that the two images are perfectly aligned
\begin{align}
I_V( \tau \Vector{x}) \sim I_U (\Vector{x}) \quad \mbox{for all pixel coordinates }\Vector{x}.
\end{align}
Here, we have chosen the standard representation of the above groups in the set of $(3 \times 3)$ real matrices,
 and the standard group action $ \tau \Vector{x}$ on the homogeneous coordinates is simply given by a matrix-vector multiplication. Note, that the inclusions $SO(2) \subset  SE(2) \subset SA(2) \subset A(2)$ hold. 
 We use the shorthand notation $\tau \circ I$ for the transformed image, i.e.
\begin{align}
(\tau \circ I) := I ( \tau \Vector{x}) \quad \mbox{for all pixel coordinates }\Vector{x}.
\end{align}

The aim of this section is to find $\tau$ by using the bimodal pair of analysis operators
$(\Matrix{\Omega}_U, \Matrix{\Omega}_V)$. The idea behind our approach is, that for an optimal transformation, the coupled sparsity measure should be minimized. Thus, we are searching for $\tau^\star \in \mathcal{G}$ such that
\begin{align}\label{optprob_register}
\tau^\star \in 
\underset{\tau \in \mathcal{G}}{\arg \min } \; g\Big(\Matrix{\Omega}_U^F I_U, \Matrix{\Omega}_V^F (\tau \circ I_V) \Big)
\end{align}

In order to tackle the above optimization problem, we follow an approach that is similar to what has been proposed in \cite{Peng2012a}. It is based on iteratively updating the estimate of $\tau$ with group elements near the identity. Locally, the Matrix exponential yields a diffeomorphism between a neighborhood of the identity in $\mathcal{G}$ and a neighborhood around $0$ in the corresponding Lie algebra $\mathfrak{g}$ of $\mathcal{G}$. For the considered Lie groups at hand, each Lie algebra is contained in 
\begin{align}
\mathfrak{g} := \left\{ \begin{bmatrix}  \Matrix{A} & \Vector{b} \\ 0 & 0 \end{bmatrix} ~|~  \Matrix{A} \in \mathbb{R}^{2 \times 2}, \Vector{b} \in \mathbb{R}^{2} \right\}  ,
\end{align} 
which is the Lie algebra of $A(2)$. Further restrictions on the parameters then lead to the corresponding Lie algebras for the sub groups. For $SO(2)$, we have $\Matrix{A}^\top = - \Matrix{A}$ and $\Vector{b}=0$, for $SE(2)$ we have $\Matrix{A}^\top = - \Matrix{A}$, and for $SA(2)$ we have $\tr \Matrix{A}=0$.

Thus, for a transformation $\delta$ which is near the identity, we have $\delta = e^\Matrix{H}$ for some matrix $\Matrix{H} \in  \mathfrak{g}$ in a neighborhood of $0$.
Now, in order to tackle the optimization problem \eqref{optprob_register} we proceed as follows.
For legibility, denote 
\begin{align}\label{eq:substitute}
F(\tau):=g\Big(\Matrix{\Omega}_U^F I_U, \Matrix{\Omega}_V^F (\tau \circ I_V) \Big).
\end{align}
We employ a geometric gradient descent method described in Section \ref{sec:line_search} on the Lie group $\mathcal{G}$ for minimizing $F(\tau)$ that updates $\tau$ in each step. 
To that end, we endow the set of $(3 \times 3)$ real matrices with the inner product

\begin{align}\label{eq:innerprod}
\langle \Matrix{H}_1, \Matrix{H}_2 \rangle_\Matrix{P}:= \tr \left((\Matrix{H}_1 \odot \Matrix{P}) \Matrix{H}_2^\top\right),
\end{align}
with $\Matrix{P}$ having positive entries and $\odot$ denoting the Hadamard product. 
The choice of $\Matrix{P}$ allows to balance the translational versus the rotational part of the chosen group, or the shearing part, respectively. This is commonly done to account for different magnitudes of the transformation parameters \cite{Klein2010}.

Choose $\tau_0 := \mathrm{id}$ as an initialization.
Then iterate the following steps until convergence.
\begin{enumerate}
\item Compute the Riemannian Gradient of $F(\delta \circ \tau)$ at $\delta = \mathrm{id}$, which is an element of the Lie algebra
\begin{align}\label{eq:gradient}
\Matrix{G}:={\rm grad}_\delta F(\delta \circ \tau)\big|_{\delta = \mathrm{id}} \in \mathfrak{g}.
\end{align}
\item Choose an approximate step-size $t^\star$ for
\begin{align}
\phi(t)=F(e^{t \Matrix{G}} \circ \tau).
\end{align}
\item Update $\tau \leftarrow  e^{t^\star \Matrix{G}}\tau$.
\end{enumerate}


For our problem at hand, we choose the Armijo rule to determine the step size. We refer to the Appendix for the derivation of the gradient of $F(\delta \circ \tau)$. 
As a stopping criterion, we choose a threshold for the norm of the Riemannian gradient.

\subsection{Evaluation}
\label{sec:alignment_experiment}
We compare our registration approach to two multimodal registration metrics, namely Mutual Information (MI) and Normalized Mutual Information (NMI) \cite{Mattes2003}. The elastix image registration toolbox \cite{Klein2010} provides the reference implementations of these metrics together with a gradient descent algorithm to find the transformation parameters. In all cases we use the standard parameters of the elastix toolbox.
In our experiments, we use intensity and depth images from the Middlebury stereo set and images from the RGB-NIR Scene Dataset \cite{Brown2011}. The RGB-NIR dataset consists of RGB images and near-infrared (NIR) images that were captured with commercial DSLR cameras using filters for the visible and infrared spectrum. The spectra do not overlap (the cutoff wavelength is about 750 nm) and the NIR images give statistically different information from the R, G and B channel.
Both datasets are very well registered and we use this registration as the ground truth and learn the operators on registered training images.

\begin{figure}
	\centering
	\begin{subfigure}[b]{0.488\columnwidth}
		\centering
		\includegraphics[width=\textwidth]{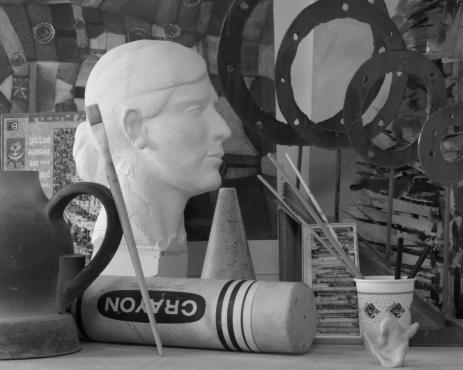}
		\caption{art intensity image}
		\label{fig:grayArt}
	\end{subfigure}%
	\,
	\begin{subfigure}[b]{0.488\columnwidth}
		\centering
		\includegraphics[width=\textwidth]{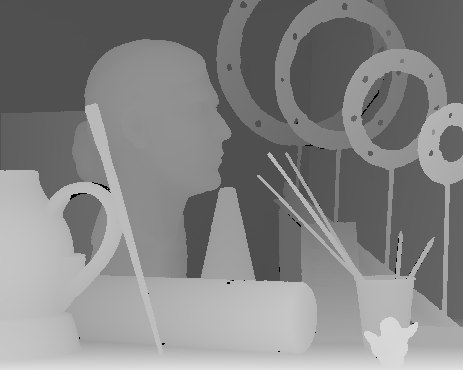}
		\caption{art depth map}
		\label{fig:depthArt}
	\end{subfigure}
	
	\medskip
	\begin{subfigure}[b]{0.488\columnwidth}
		\centering
		\includegraphics[width=\textwidth]{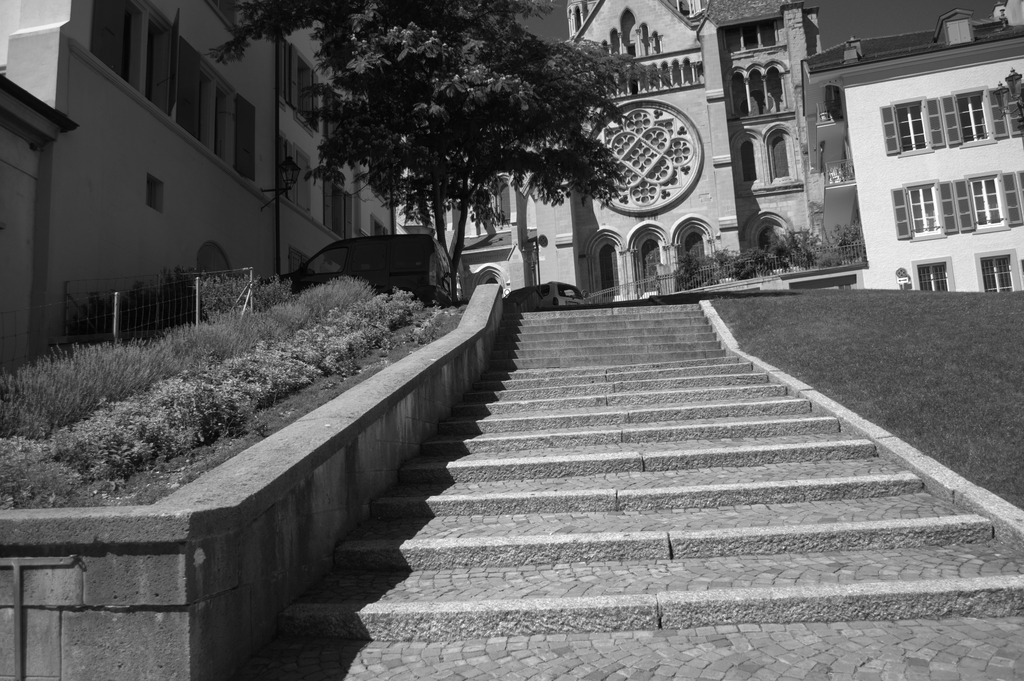}
		\caption{intensity old building}
		\label{fig:rgbOld}
	\end{subfigure}%
	\,
	\begin{subfigure}[b]{0.488\columnwidth}
		\centering
		\includegraphics[width=\textwidth]{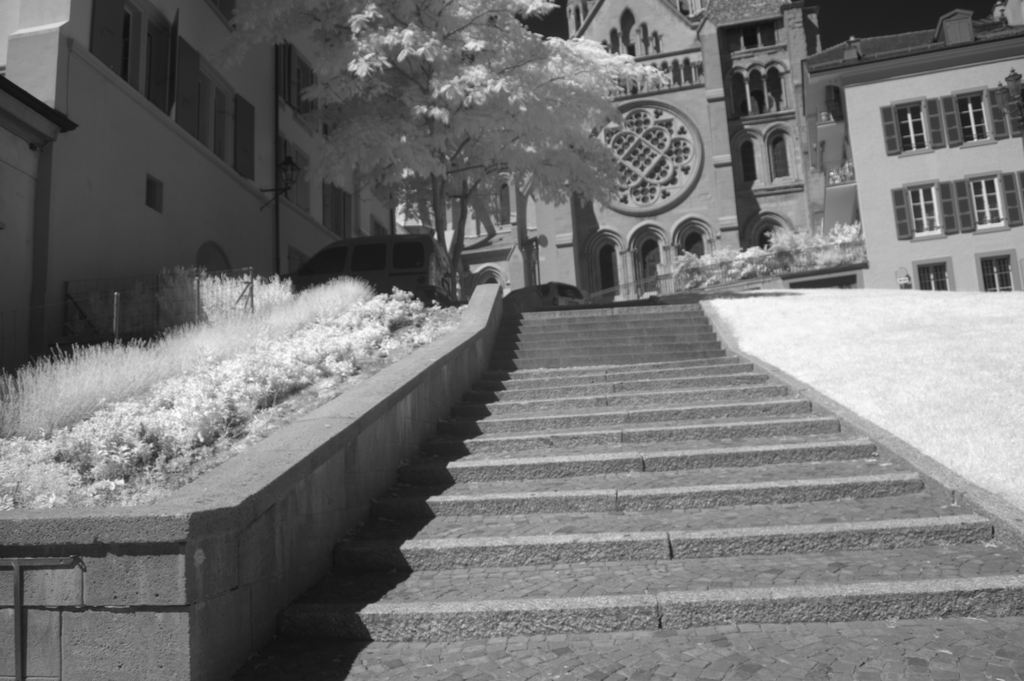}
		\caption{NIR old building}
		\label{fig:nirOld}
	\end{subfigure}
	
	\medskip
	\begin{subfigure}[b]{0.488\columnwidth}
		\centering
		\includegraphics[width=\textwidth]{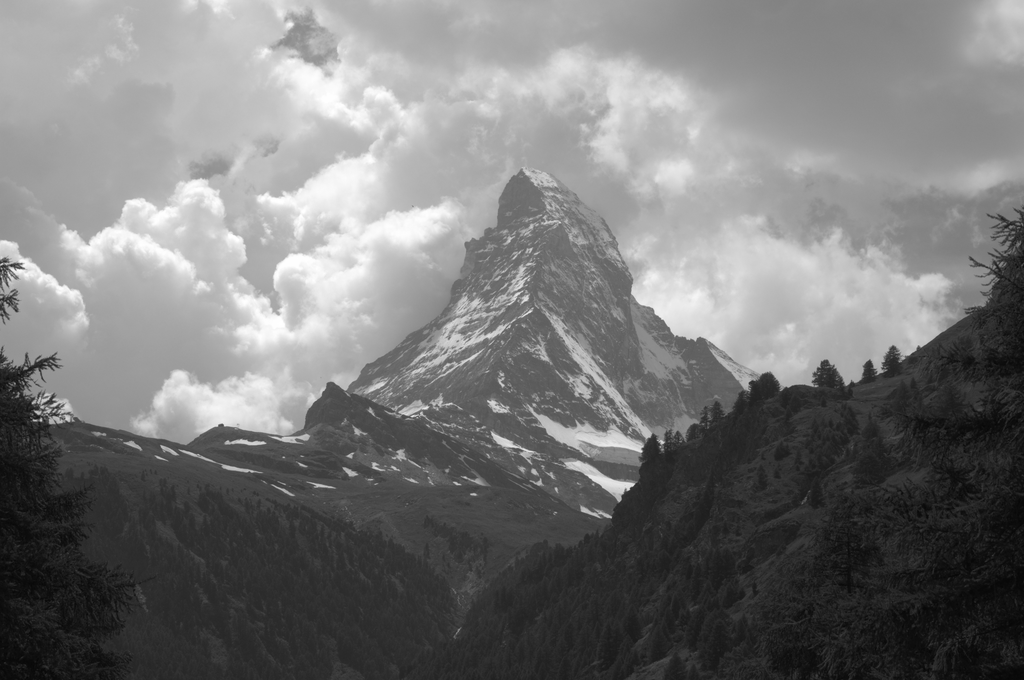}
		\caption{intensity mountain}
		\label{fig:rgbMountain}
	\end{subfigure}%
	\,
	\begin{subfigure}[b]{0.488\columnwidth}
		\centering
		\includegraphics[width=\textwidth]{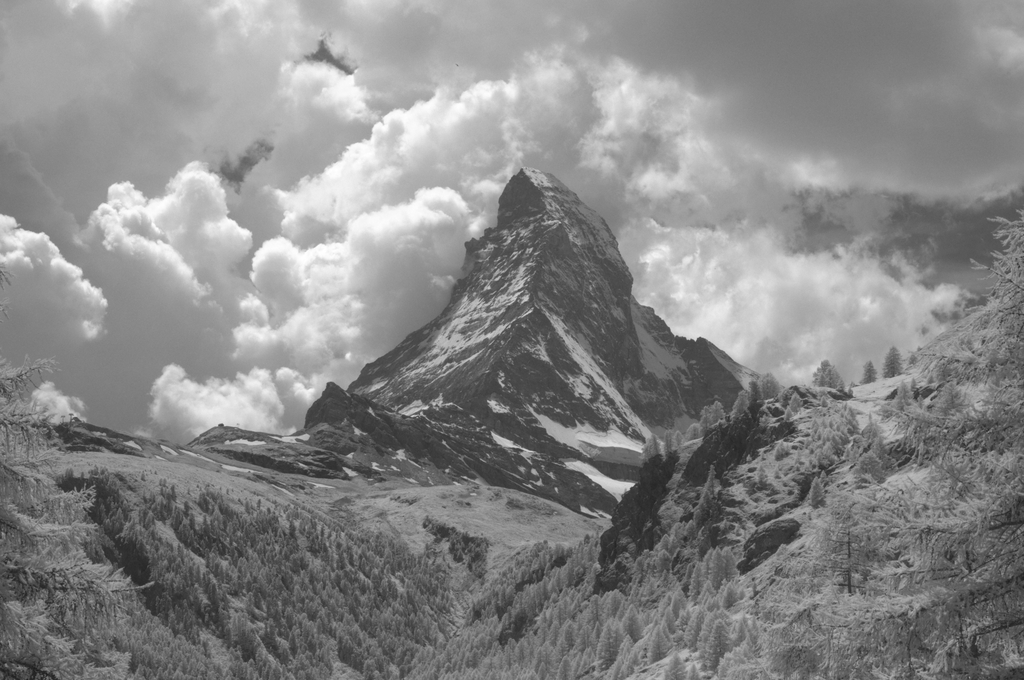}
		\caption{NIR mountain}
		\label{fig:nirMountain}
	\end{subfigure}
	\caption{Example images used in the registration experiments. The intensity and depth image pair differ significantly and is challenging for multimodal registration algorithms. The NIR images are more similar to the intensity images and differ mainly in areas with vegetation and sky.}\label{fig:rgbnir}
\end{figure}

We train one fixed operator pair for each of the registration scenarios intensity+depth and intensity+NIR. For the intensity and depth setup, we use the same operator as in the reconstruction experiments in Section \ref{sec:middlebury_evaluation}. For the intensity and NIR setup, we followed the same learning procedure, randomly collecting $M=15000$ pairs of square sample patches of size $\sqrt{n}=5$ from a total of $9$ images in the training set, one from each category which we then exclude from testing. We set the learning parameters to $\nu=200, \kappa_I=250, \kappa_N=1000, \mu_I=250$ and $\mu_N=1000$. All other parameters are the same as for intensity and depth.

In order to evaluate the result of the registration of one image pair, we apply a synthetic deregistration to one of the images. This deregistration consists of a translation and a rotation and subsequently the registration algorithm searches for a transformation that belongs to the special Euclidian group. Both the elastix toolbox and our algorithm work on a Gaussian image pyramid of four levels.

\begin{table*}[t]
	  \centering
	  \resizebox{\linewidth}{!}{
		  \begin{tabular}{rrccc}
		    \toprule
					deregistration (x,y,$\theta$) & method & intensity-depth (art) & intensity-NIR (old building) & intensity-NIR (mountain) \\
		    \midrule
\multicolumn{1}{c}{\multirow{3}[0]{*}{0, 0, 10}} & MI & 14.70, 50.14, -2.01 & 6.14, 1.56, -3.01 & \textbf{0.16}, \textbf{-0.85}, \textbf{-3.52}\\
\multicolumn{1}{c}{} & NMI & 9.42, 18.66, -7.83 & 2.49, 4.26, -9.17 & -1.43, -1.05, -3.85\\
\multicolumn{1}{c}{} & our method & \textbf{-1.11}, \textbf{-1.41}, \textbf{0.11} & \textbf{0.35}, \textbf{0.22}, \textbf{0.01} & -0.34, -2.95, -8.31\\
\midrule
\multicolumn{1}{c}{\multirow{3}[0]{*}{-5, -5, 0}} & MI & -1.06, \textbf{1.13}, 0.02 & -0.21, -0.14, 0.05 & 0.10, -0.06, \textbf{0.05}\\
\multicolumn{1}{c}{} & NMI & 7.02, 4.96, \textbf{0.01} & \textbf{0.03}, \textbf{0.10}, \textbf{0.02} & \textbf{0.05}, \textbf{0.01}, 0.06\\
\multicolumn{1}{c}{} & our method & \textbf{-1.00}, -1.13, 0.03 & -1.03, 2.34, 2.72 & 2.56, 0.41, 0.06\\
\midrule
\multicolumn{1}{c}{\multirow{3}[0]{*}{10, 0, 5}} & MI & 8.60, 18.71, -2.49 & -8.59, 2.26, -1.32 & 2.64, 0.08, -1.76\\
\multicolumn{1}{c}{} & NMI & 3.44, 9.79, -3.27 & -8.22, 2.27, -1.35 & 2.58, \textbf{0.05}, -1.71\\
\multicolumn{1}{c}{} & our method & \textbf{-0.90}, \textbf{-0.81}, \textbf{0.19} & \textbf{0.02}, \textbf{0.23}, \textbf{0.06} & \textbf{0.61}, 0.37, \textbf{-0.02}\\
\midrule
\multicolumn{1}{c}{\multirow{3}[0]{*}{-5, -5, 5}} & MI & 1.38, 11.12, -2.65 & -7.53, 1.79, -1.43 & 1.57, -0.23, -1.77\\
\multicolumn{1}{c}{} & NMI & 3.04, 8.87, -3.01 & -8.83, 1.86, -1.36 & 1.58, -0.27, -1.73\\
\multicolumn{1}{c}{} & our method & \textbf{-0.22}, \textbf{-1.40}, \textbf{0.28} & \textbf{0.22}, \textbf{-0.13}, \textbf{0.14} & \textbf{0.82}, \textbf{-0.01}, \textbf{-0.28}\\
		    \bottomrule
		  \end{tabular}%
	  }
	\caption{Registration residual for different synthetic translations and rotations. Values for the translation in x and y direction are given in pixels, the angle $\theta$ is given in degrees.}
	\label{tab:registration_results}%
\end{table*}%

\begin{figure*}[hbtp]
\centering
\includegraphics[width=\textwidth]{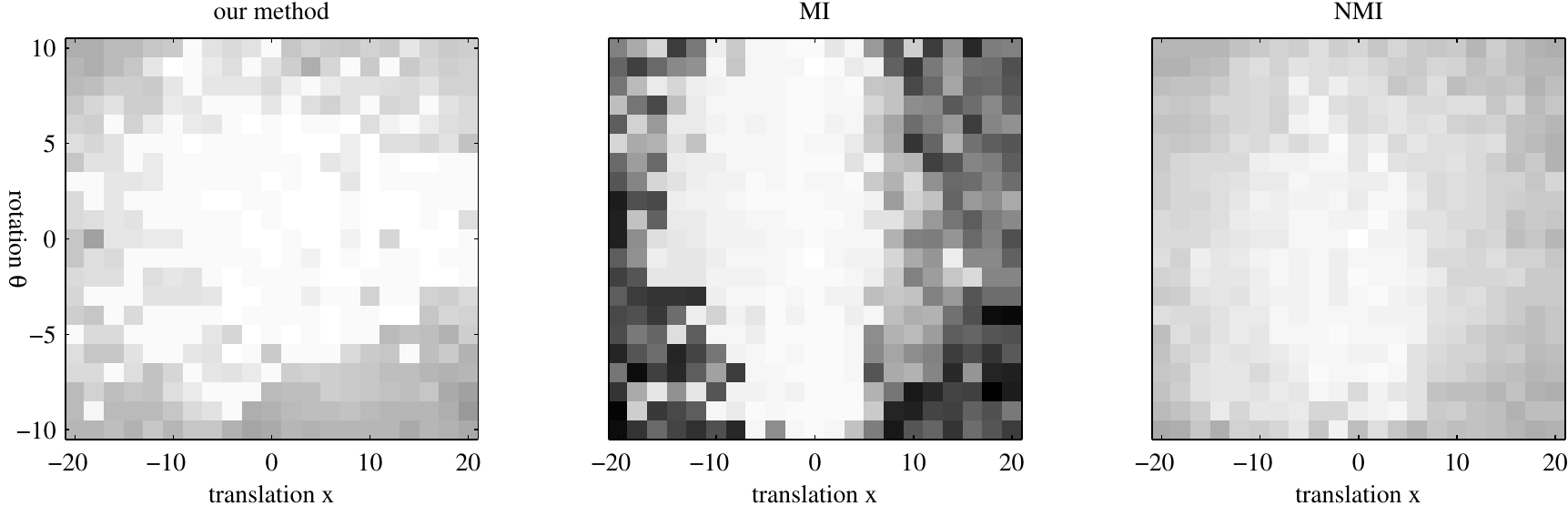}
\caption{Remaining combined registration error for different initial deregistrations which consist of a translation in x-direction and a rotation $\theta$. White and black areas correspond to small and large errors, respectively. MI fails to register the images for large translations. Our method achieves the smallest remaining error and can handle large translations and rotations.}
\label{fig:remain_reg_error}
\end{figure*}

\begin{figure}
	\centering
	\begin{subfigure}[b]{0.488\columnwidth}
		\centering
		\includegraphics[width=\textwidth]{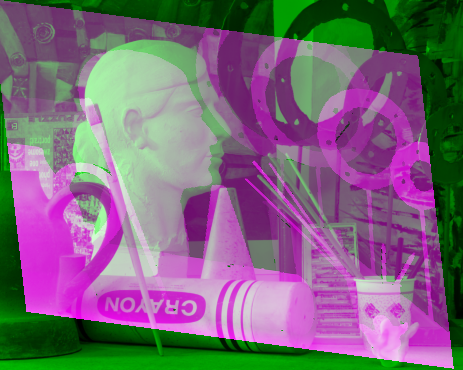}
		\caption{deregistered input}
		\label{fig:affineBefore}
	\end{subfigure}%
	\,
	\begin{subfigure}[b]{0.488\columnwidth}
		\centering
		\includegraphics[width=\textwidth]{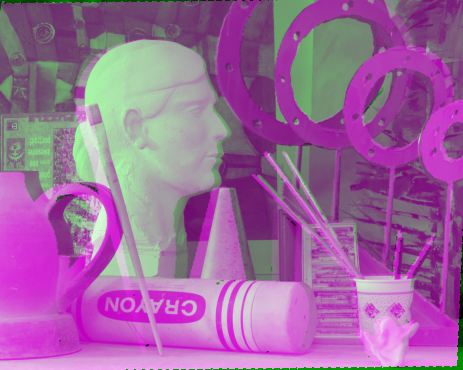}
		\caption{after registration}
		\label{fig:affineReg}
	\end{subfigure}
	\caption{Example of an intensity and depth image pair before and after the registration process using an affine transformation.}\label{fig:affine}
\end{figure}

Table \ref{tab:registration_results} shows the remaining registration error after running the different registration algorithms. Our method achieves comparable or better results than MI or NMI for all of the modalities. The MI and NMI algorithms fail to register the intensity and depth image pair in most of the cases but achieve better results in the intensity-NIR images. This can be explained by the fact that intensity and depth are much less alike than intensity and near-infrared (see Figure \ref{fig:rgbnir}). 
Figure \ref{fig:remain_reg_error} shows the remaining registration error for various initial deregistrations of the intensity and depth image pair. We define the remaining combined registration error as
\begin{equation}
\epsilon = \sqrt{\epsilon_x^2 + \epsilon_y^2 + \epsilon_{\theta}^2},
\end{equation}
where $\epsilon_x$ and $\epsilon_y$ denote the remaining translation error in x- and y-direction (in pixels) and $\epsilon_{\theta}$ denotes the remaining rotation error (in degrees).
White areas in Figure \ref{fig:remain_reg_error} correspond to small registration errors ($\epsilon < 1$) and black areas show large errors ($\epsilon > 50$) where the registration has failed.
It can be seen that MI is susceptible to large translations and fails to align the images correctly. The direct comparison of our method and NMI shows that both algorithms can handle the initial deregistration better than MI but our method achieves smaller remaining errors over a wider range of deregistration values.

\section{Conclusion and Discussion}
\label{sec:conclusion}
We have introduced a way to model the interdependencies of two image modalities by extending the co-sparse analysis model. The coupled analysis operators are learned by minimizing a coupled sparsity function via a conjugate gradient method on an appropriate manifold. The manifold setting allows to constrain the rows of the operators a priori to zero mean and unit norm, which accounts for the intuition that contrast in image modalities is most informative. 

We evaluate the descriptive power of the presented model in two application scenarios. First, we have used it as a regularizer for inverse problems in imaging and provided numerical experiments for depth map super-resolution, when a high resolution intensity image is available for the same scene. 
As a second application scenario, we have considered the problem of bimodal image registration. An algorithm on Lie groups has been proposed that uses a previously learned bimodal model to register intensity and depth images as well as intensity and near infrared (NIR) images. 

Despite these convincing results, using our model for bimodal image processing certainly has some limitations.
The model is based on local assumptions and thus fails to regularize tasks where large areas of one modality are not available. For example, inpainting of large gaps in the image typically fails due to the lack of global support, as is also reported in \cite{kiechle2013jid}.

Now clearly, the introduction of a new model poses at least as many questions as it does provide answers.
Our evaluation shows promising results for intensity/depth and intensity/NIR images. 
The question how our model will perform for other modality compositions remains open at this point.
For example, it would be interesting to investigate its applicability in medical imaging, where different modalities like MRI or PET play an important role.
%
We leave these
questions to future work.

%
%


\appendix
\section{Appendix}
\label{sec:appendix}
\subsection{Derivation of the Riemannian gradient in Section \ref{sec:alignment}}

In this section we derive the Riemannian gradient in Eq. \eqref{eq:gradient} for the bimodal alignment algorithm.
We make use of the following criterion for its derivation. Let $\langle \cdot,\cdot \rangle_\Matrix{P}$ be the Riemannian metric on the Lie group $\mathcal{G}$ inherited from \eqref{eq:innerprod} and let $F(\cdot)$ be a smooth real valued function on $\mathcal{G}$. Then the Riemannian gradient of $F$ at $\delta \in \mathcal{G}$ is the unique vector $\Matrix{G} \in T_\delta \mathcal{G}$, the tangent space at $\delta$, such that 
\begin{align}\label{eq:grad_comp}
\frac{{\rm d} }{{\rm d} t } \Big|_{t=0} F(e^{t \Matrix H}\delta) = \langle \Matrix{H},\Matrix{G} \rangle_\Matrix{P} 
\end{align}
holds for all tangent elements $\Matrix{H} \in T_\delta \mathcal{G}$.

For our purpose, we compute the gradient at $\delta = {\rm id}$.
Now let $B$ be the image region in which we want to align the modalities $I_U$ and $I_V$. We assume that $B$ is rectangular and denote by
\begin{align}
I(\Vector{x})_{\Vector{x} \in B}
\end{align}
the vectorized version of $I$ over the domain $B$.
Using Equation \eqref{eq:substitute} and the fact that $\Vector{c}:= \Matrix{\Omega}^F_U I_U$ is a constant vector, we compute by the chain rule that

\begin{align}
& \tfrac{{\rm d} }{{\rm d} t } \Big|_{t=0}  F(e^{t \Matrix H}\tau) = \tfrac{{\rm d} }{{\rm d} t } \Big|_{t=0}  g\left(\Vector{c}, \Matrix{\Omega}^F_V \left[(e^{t \Matrix H}\tau) \circ I_V \right]\right) \notag \\
& = \nabla g\left(\Vector{c}, \Matrix{\Omega}^F_V I_V (\tau \Vector{x})_{\Vector{x} \in B} \right)^\top
\Matrix{\Omega}^F_V \left[ \tfrac{{\rm d} }{{\rm d} t } \Big|_{t=0} I_V (e^{t \Matrix{H}} \tau \Vector{x})_{\Vector{x} \in B}\right].
\end{align}
The last bracket is a vector where each of its entries is computed as
\begin{align}
& \tfrac{{\rm d} }{{\rm d} t } \Big|_{t=0} I_V (e^{t \Matrix{H}}\tau \Vector{x}) = \nabla I_V(\tau \Vector{x})^\top \Matrix{H} \tau \Vector{x} = {\rm vec} (\tau \Vector{x} \otimes \nabla I_V(\tau \Vector{x}))^\top {\rm vec}(\Matrix{H}),
\end{align}
where as usual, ${\rm vec}(\cdot)$ denotes the linear operator that stacks the columns of a matrix among each other and $\otimes$ is the Kronecker product.
Note, that since we stick to the representation with homogeneous coordinates, $\nabla I_V(\Vector{x}) \in \mathbb{R}^3$ is the common image gradient of $I_V$ with an additional $0$ in the third component. 

\noindent
Thus with
\begin{align}
&\Vector{r}^\top\colon\!{=} \nabla g\left(\Vector{c}, \Matrix{\Omega}_V^F I_V (\tau \Vector{x})_{\Vector{x} \in B} \right)^\top
\Matrix{\Omega}_V^F \left( {\rm vec} (\tau \Vector{x} \otimes \nabla I_V(\tau \Vector{x}))^\top    \right)_{\Vector{x} \in B},
\end{align}
we have 
\begin{align}
\frac{{\rm d} }{{\rm d} t } \Big|_{t=0} F(e^{t \Matrix H}\delta) = \Vector{r}^\top {\rm vec}(\Matrix{H}) & = \tr({\rm vec}^{-1}(\Vector{r})  \Matrix{H}^\top) = \langle {\rm vec}^{-1}(\Vector{r})\odot \hat{\Matrix{P}} , \Matrix{H} \rangle_{\Matrix{P}},
\end{align}
where the entries of $\hat{\Matrix{P}}$ are the inverse of the entries of $\Matrix{P}$.

Using Equation \eqref{eq:grad_comp}, the Riemannian gradient is therefore the orthogonal projection of ${\rm vec}^{-1}(\Vector{r})\odot \hat{\Matrix{P}}$ with respect to $\langle \cdot, \cdot \rangle_{\Matrix{P}}$ onto the tangent space of $\delta = {\rm id}$, which is nothing else than the Lie algebra $\mathfrak{g}$, i.e.
\begin{align}
{\rm grad}_\delta F(\delta \circ \tau) = \mathrm{\Pi}_{\mathfrak g}\left( {\rm vec}^{-1}(\Vector{r})\odot \hat{\Matrix{P}} \right). 
\end{align}

\noindent
If we further assume for the entries $p_{ij}$ of $\Matrix{P}$ that 
\begin{align}
p_{11}=p_{22} \text{ and } p_{12}=p_{21},
\end{align} 
then, for the considered Lie groups, these projections are explicitly given by
\begin{align}
\mathrm{\Pi}_{SO}(\Matrix{X})& = \begin{bmatrix} \tfrac{1}{2}(\Matrix{X}_{11}-\Matrix{X}_{11}^\top) & 0 \\ 0 & 0 \end{bmatrix} \\
\mathrm{\Pi}_{SE}(\Matrix{X})& =  \begin{bmatrix} \tfrac{1}{2}(\Matrix{X}_{11}-\Matrix{X}_{11}^\top) & \Vector{x}_{12} \\ 0 & 0 \end{bmatrix} \\
\mathrm{\Pi}_{SA}(\Matrix{X})& =  \begin{bmatrix} (\Matrix{X}_{11}-\tfrac{1}{2}\tr(\Matrix{X}_{11}) \Matrix{I}_2) & \Vector{x}_{12} \\ 0 & 0 \end{bmatrix}\\
\mathrm{\Pi}_{A}(\Matrix{X}) & = \begin{bmatrix} \Matrix{X}_{11} & \Vector{x}_{12} \\ 0 & 0 \end{bmatrix},
\end{align}
where $\Matrix{X} \in \mathbb{R}^{3 \times 3}$ is partitioned as
\begin{align}
\Matrix{X} = \begin{bmatrix} \Matrix{X}_{11} & \Vector{x}_{12} \\ \Vector{x}_{21}^\top & x_{22} \end{bmatrix}.
\end{align}


\bibliographystyle{spmpsci}      
\bibliography{PaperCitations}   


\end{document}